\documentclass[runningheads,a4paper]{llncs}

\usepackage{amssymb}
\usepackage{graphicx}
\usepackage{url}
\usepackage{apacite}
\usepackage[toc,page]{appendix}
\newcommand{\keywords}[1]{\par\addvspace\baselineskip
\noindent\keywordname\enspace\ignorespaces#1}
\begin{document}

\mainmatter  

%
\title{Emotional Musical Prosody: Validated Vocal Dataset for Human Robot Interaction}

\author{Richard Savery, Lisa Zahray, Gil Weinberg \thanks{This material is based upon work supported by the National Science Foundation under Grant No. 1925178 
}}
%
\authorrunning{Savery, Zahray, Weinberg}

\titlerunning{{\em 2020 Joint Conference on AI Music Creativity}, Demo Paper}

\institute{Georgia Tech Center for Music Technology\\ \email{rsavery3@gatech.edu}}

%


\maketitle

\begin{abstract}
Human collaboration with robotics is dependant on the development of a relationship between human and robot, without which performance and utilization can decrease. Emotion and personality conveyance has been shown to enhance robotic collaborations, with improved human-robot relationships and increased trust. One under-explored way for an artificial agent to convey emotions is through non-linguistic musical prosody. In this work we present a new 4.2 hour dataset of improvised emotional vocal phrases based on the Geneva Emotion Wheel. This dataset has been validated through extensive listening tests and shows promising preliminary results for use in generative systems.


\keywords{robotics, emotion, prosody, music}
\end{abstract}

\section{Introduction}
As the use of robotics and artificial agents continue to expand, there is a growing need for better forms of communication between human and computer. While speech based mediums are common in home assistants and robots, we contend that for many human-agent collaborations, semantic meaning is not always required. One alternate to speech communication is non-linguistic emotional musical prosody where a robot communicates through musical phrases. Using musical phrases can avoid the challenges of uncanny valley, allowing robot interaction to have it's own human-inspired form of communication \cite{savery_finding_2019}. In past work we have successfully shown the capabilities of prosody in robotic systems through simulations of a potential generative system. This includes increased trust in social robots \cite{savery2019establishing} and industrial robots \cite{savery2020emotional}. In this paper we present the collection and validation of a emotional prosody dataset that can be used to improve future human robot interactions.

\section{Custom Dataset}
\begin{figure}[t]
\centering
\begin{minipage}{.5\textwidth}
  \centering
  \includegraphics[width=1\linewidth]{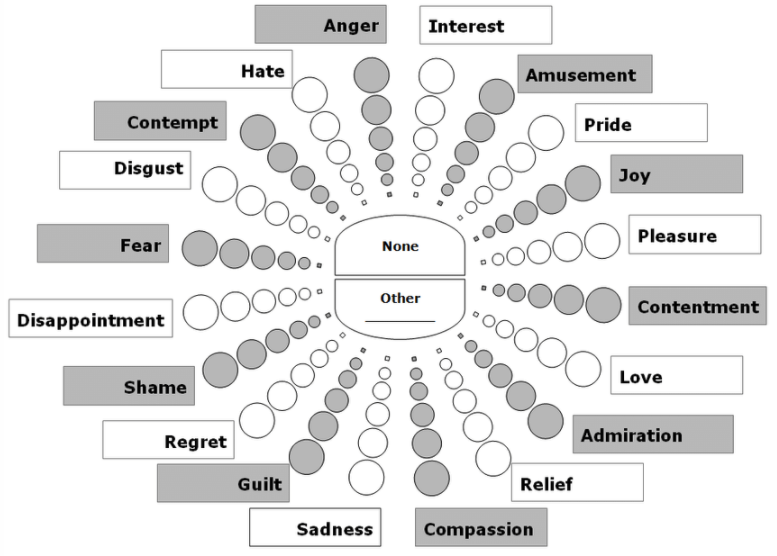}
  \caption{Geneva Emotion Wheel}
  \label{fig:geneva}
\end{minipage}%
\begin{minipage}{.5\textwidth}
  \centering
  \includegraphics[width=0.8\linewidth]{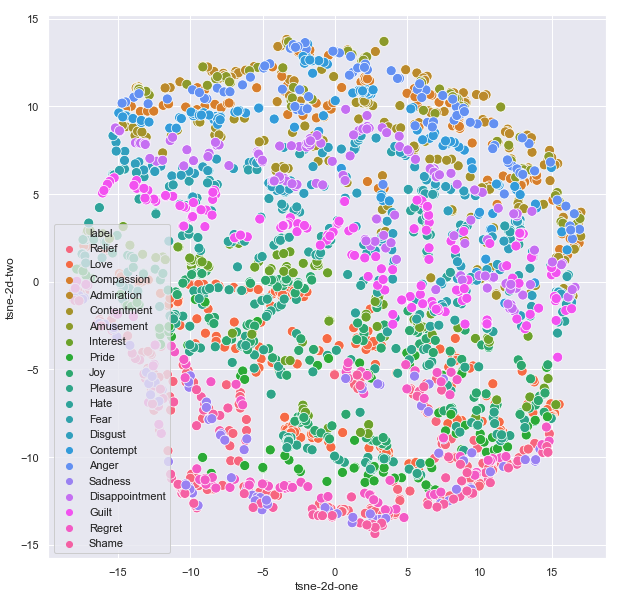}
  \caption{Vanilla VAE Latent Space}
  \label{fig:test2}
\end{minipage}
\end{figure}

Our custom dataset with 4.2 hours of audio was created with the singer Mary Carter. After a pilot study collecting data, we decided to use the Geneva Emotion Wheel (GEW) \cite{sacharin2012geneva}, which is a circular model, containing 20 emotions with position corresponding to the circumplex model (see Figure \ref{fig:geneva}).

One of the primary advantages of the GEW is that it includes 20 different emotions, but these emotions can also be reduced to four separate classes which align with a quadrant from the circumplex model. GEW also includes most of the Eckman's basic emotions - fear, anger, disgust, sadness, happiness - only leaving out surprise. The ability to use different models of emotion allows for significant future use cases of the dataset \cite{saverysurvey}.

This dataset only has one musician, and therefore only captures one person's perspective on musical emotion. While the dataset can make no claim to represent all emotion and does not create a generalized emotion model, we believe using one person has advantages. By having only one vocalist generative systems have the possibility of recreating one person's emotional style, instead of incorrectly aggregating multiple peoples to remove distinctive individual and stylistic features.

\subsection{Process and Data}
Carter was paid \$500 and recorded the samples over a week long period at her home studio, using a template we created in Logic while maintaining the same microphone positioning. For the samples we requested phrases to be between 1 and 20 seconds, and to spend about 15 minutes on each emotion, allowing jumping between any order of the emotions. We allowed deletion of a phrase if she felt after singing it did not fit the emotion. The final recorded dataset includes 2441 phrases equalling 4.22 hours of data with an average of 122 for each emotion. Samples from the dataset can be heard online.\footnote{www.richardsavery.com/prosodycvae}

\subsection{Dataset Validation}\label{sec:dataval} To validate the dataset, we performed a study with 45 participants from Prolific and Mechanical Turk, paying each \$3. Each question in the survey asked the participant to listen to a phrase and select a location on the wheel corresponding to the emotion and intensity they believed the phrase was trying to convey. Phrases fell under two categories of ``best'' and ``all'', with each participant seeing 60 total phrases selected at random. The ``best'' category consisted of 5 phrases for each emotion that were hand-selected as best representing that emotion, ensuring an even distribution of phrase lengths in each emotion set. The ``all'' category consisted of a phrase sampled from all phrases in the dataset for that emotion, with a new phrase randomly selected for each participant. Rose plots of the validation results that combine the ``best'' and ``all'' categories can be seen in the appendix, separated into each Geneva Wheel quadrant.



\subsection{Dataset to Midi}\label{sec:datasettomidi}
We converted each phrase's audio into a midi representation to use as training data. We first ran the monophonic pitch detection algorithm CREPE \cite{kim2018crepe} on each phrase, which outputs a frequency and a confidence value for a pitch being present every 0.01 seconds. As the phrases included breaths and silence, it was necessary to filter out pitches detected with low confidence. We applied a threshold followed by a median filter to the confidence values, and then forced each detected pitch region to be at least 0.04 seconds long. We then converted the frequencies to midi pitches. We found the most common pitch deviation for each phrase using a histogram of deviations, shifting the midi pitches by this deviation to tune each phrase. 

\section{Conclusion}
We have so far analyzed the dataset by training a Vanilla VAE (without emotion labels) on the pitch alone. Figure \ref{fig:test2} shows the latent space reduced from 100 dimensions through t-distributed stochastic neighbor embedding (t-SNE) \cite{hinton2003stochastic}. This demonstrates that the latent space is able to separate by emotion, without any knowledge of the emotional labels. We then used a conditional VAE (conditioned on emotion labels) to generate phrases with playback through a separate audio sampler. These phrases were validated following the format of our dataset evaluation, and the results are shown in the appendix.
In the future we will run extensive studies with generated prosody applied to human-robot interactions. This will take place between varying group sizes from one human and robot, to groups of humans and robots with different embedded personalities. We expect for emotional musical prosody to enable many future collaborations between human and robot.

\bibliographystyle{apacite}
\bibliography{CSMC_MUME_LaTeX_Template.bib}

\section{Appendix}

\begin{figure}[]
	\centering
	\includegraphics[height=6.4cm]{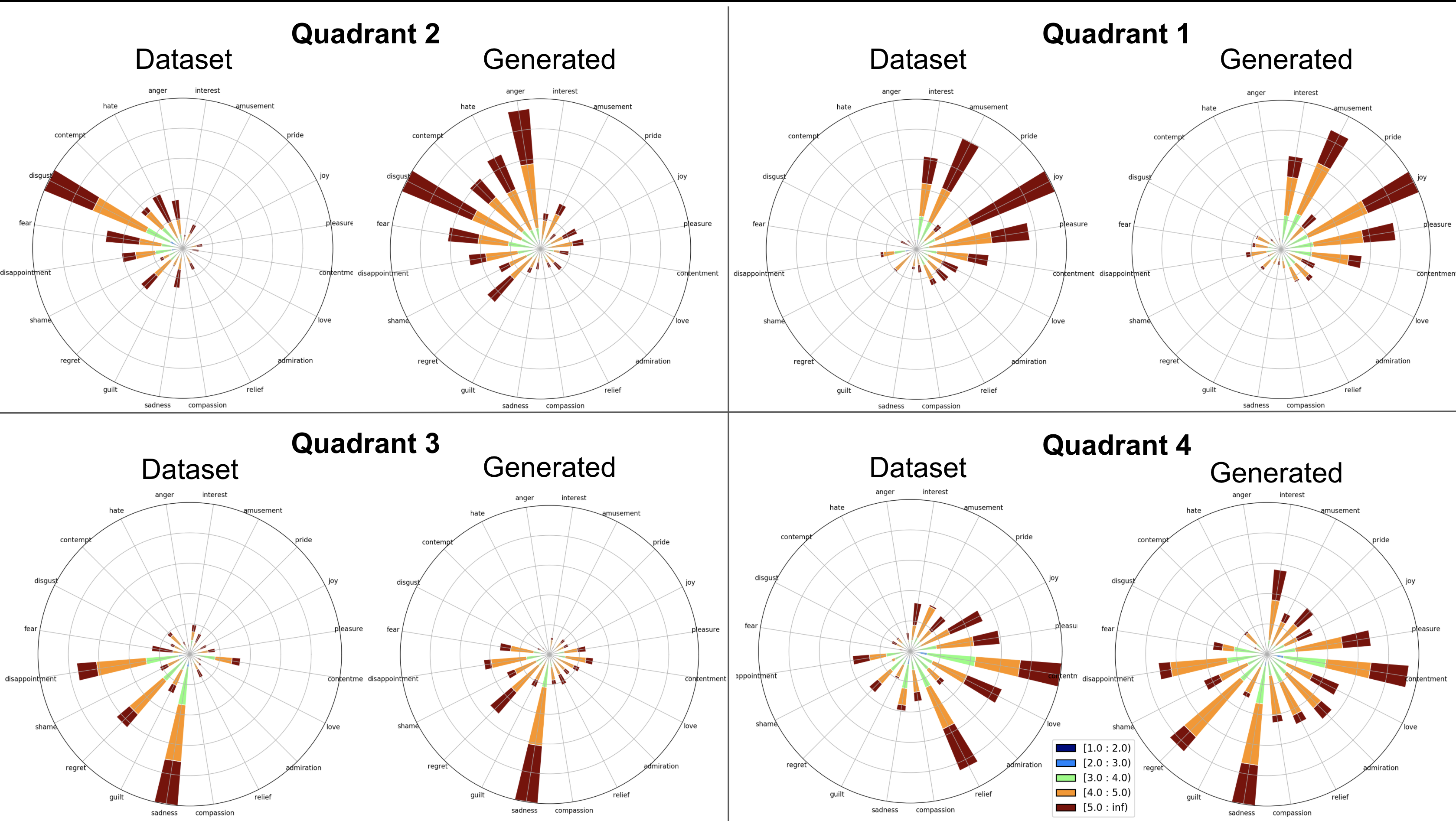}
	\caption{Rose plots of dataset validation for each emotion quadrant}
	\label{fig:roseplots}
\end{figure}
\end{document}